\documentclass[twoside,11pt]{article}

\usepackage{dnd}
\usepackage[T1]{fontenc}
\usepackage{times}
\usepackage{linguex}
\usepackage{xcolor}
\usepackage{mathtools}
\usepackage{multirow}
\usepackage{amsmath}
\usepackage[bottom]{footmisc}
\usepackage{array}
\usepackage{lipsum}
\usepackage{CJKutf8}
\usepackage{arabtex}
\usepackage{utf8}
\usepackage{accents}
\usepackage{booktabs}
\usepackage{appendix}

\newcommand*\underdot[1]{%
  $\underaccent{\dot}{#1}$}
  
  \usepackage{accents}
\newcommand{\ubar}[1]{$\underaccent{\bar}{#1}$}

\def\endabstract{\egroup}

\setcode{utf8}

\newcommand{\chinese}[1]{\begin{CJK}{UTF8}{gbsn}#1\end{CJK}}

\newcommand\blfootnote[1]{%
  \begingroup
  \renewcommand\thefootnote{}\footnote{#1}%
  \addtocounter{footnote}{-1}%
  \endgroup
}

\dndheading{xx(y)}{2021}{xx--yy}{Amir Zeldes}{10.5087/dad.2021.xxx}

\ShortHeadings{Can we Fix the Scope for Coreference?}{Zeldes}

\begin{document}

\title{Opinion Piece: Can we Fix the Scope for Coreference? \\Problems and Solutions for Benchmarks beyond OntoNotes}

\author{\name Amir Zeldes \email amir.zeldes@georgetown.edu \\
       \addr Georgetown University\\
       Department of Linguistics
       }


\maketitle{}

\begin{abstract}

Current work on automatic coreference resolution has focused on the OntoNotes benchmark dataset, due to both its size and consistency. However many aspects of the OntoNotes annotation scheme are not well understood by NLP practitioners, including the treatment of generic NPs, noun modifiers, indefinite anaphora, predication and more. These often lead to counterintuitive claims, results and system behaviors. This opinion piece aims to highlight some of the problems with the OntoNotes rendition of coreference, and to propose a way forward relying on three principles: 1.~a focus on semantics, not morphosyntax; 2.~cross-linguistic generalizability; and 3.~a separation of identity and scope, which can resolve old problems involving temporal and modal domain consistency.

\end{abstract}


\section{Introduction} \label{introduction}

\blfootnote{I would like to thank Anna Nedoluzhko, Massimo Poesio, Sameer Pradhan, Nathan Schneider and the anonymous reviewers for valuable comments and discussions about previous versions of this paper; the usual disclaimers apply.}

Coreference resolution is the task of delineating and grouping together referring expressions in text so that spans of text referring to the same discourse entity are clustered together. The past decade has seen remarkable improvements in the consistency and performance of coreference resolution, first through the creation of large ($>$1 M tokens) benchmark data in OntoNotes (\citealt{HovyMarcusPalmerEtAl2006,WeischedelPradhanRamshawEtAl2012}; hence \textsc{ON}), and then through use of that dataset in the CoNLL shared task on coreference resolution \citep{PradhanRamshawMarcusEtAl2011} and the development of evaluation metrics (\citealt{PradhanEtAl2014}; see \citealt{MoosaviStrube2016} for criticism). With the advent of end-to-end neural approaches to coreference resolution \citep{LeeHeLewisEtAl2017}, we have seen scores on coreference resolution rise from the mid-50s at the 2011 shared task to current SOTA scores around 80 points when gold speaker information is provided \citep{JoshiEtAl2019, WuEtAl2020}, as evaluated by the standard metrics on the ON test set.

At the same time, coreference resolution using the ON scheme as a target has been plagued by a number of issues: the lack of annotation of singletons (entities mentioned only once) has led to systems conflating referentiality recognition (whether an expression in fact refers to some entity) with coreference recognition (whether a referring expression is mentioned more than once, see \citealt{LeeHeLewisEtAl2017,ZhangEtAl2018}); omission of annotation of indefinite anaphors has led to counterintuitive system behaviors (e.g.~repeated mention of specific indefinites and generics is ignored; see Section \ref{sec:indef} and the Appendix for examples); and idiosyncratic handling of some constructions (especially copula predication, compounding) has meant that even when systems perform well in predicting ON-like annotations, the results are often less informative than they could be for practical applications (see Section \ref{sec:pred}). Additionally, because the ON guidelines for English depend on morphosyntactic constructs such as indefinite NPs, copula predication and compounding, they are not readily applicable to other languages, leading to rather different practices in corpora for languages other than English, including in the Arabic and Chinese sections of the OntoNotes corpus itself.

I would like to start by saying that I have been a passionate admirer and intensive user of OntoNotes for a long time, but I also suspect that dissatisfaction with some aspects of its annotation scheme is widespread: this is evidenced by the emergence of the Coreference Resolution Beyond OntoNotes workshops since 2016 (CORBON, and later CRAC, `Computational Models of Reference,~Anaphora and Coreference'), the continuing creation of resources with different annotation schemes (cf.~\citealt{StoyanovEtAl2009,ZeldesZhang2016}; see \citealt{PoesioPradhanRecasensEtAl2016} for an overview), the focus of the 2021 coreference shared task on non-ON data, and most recently the launching of a Universal Anaphora (UA) initiative, aimed at bringing some consistency into the growing variety of divergences between resources.\footnote{See \url{http://universalanaphora.org/}} However, I suspect that while we as a research community can arrive at a consensus about this dissatisfaction, it will be much harder to reach a consensus about what a solution might look like, and how it could be attained feasibly. My goal in writing this overview is therefore first to outline some of the reasons why we need to do something, and my audience is, in the first instance, the community working on language resources for coreference, and in the second instance, NLP researchers, and especially grad students, who are entering the field and may get the impression that the ON standard represents a final end point of what we want from coreference resolution. The suggestions below probably form a continuum between more and less controversial, and their ranking on different researchers' wish lists will differ as well based on their respective goals, but it's a discussion we can only have if the commonalities and differences between coreference datasets and task definitions are laid out clearly.\footnote{For readers who prefer to start with worked examples, see the Appendix for some analyses contrasting the ON view of coreference with a more unrestricted annotation scheme.}

\section{The 90\% and 10\% in OntoNotes}

The development and release of ON were landmark achievements for a number of annotation types covered by the corpus, but especially transformative for coreference resolution. One of the declared goals of the project, as presented in the \citeauthor{HovyMarcusPalmerEtAl2006} paper and its title, was to reach a 90\% solution: agreement scores for annotations had to be in the 90s, casting consistency as a necessary condition for any guidelines developed for the corpus. For coreference, this meant focusing on some of the easier phenomena to agree on, such as antecedents for pronominal anaphora, while more tricky cases, such as indefinite anaphors, were left unannotated -- not because they were deemed disinteresting or unuseful, but because they were a lower priority target (Sameer Pradhan, p.c.). This decision led to high agreement, but also to a number of unexpected results. 

In this section I will survey some of the most surprising cases for lay users or researchers new to coreference resolution; for quite a few cases, I am certain that the majority of people using off-the-shelf coreference tools are not well aware of the specifics or consequences, and many people working with ON data directly are also not likely to know about all. In the following section I will argue that for all of these, semantic reference to the same entity, rather than morphosyntactic environments, should guide the coreference decision.

\subsection{Indefinites and generics} \label{sec:indef}

ON guidelines prohibit ``generic, underspecified, and abstract nominal mentions'' \citep[5]{BBNONEngGuidelines} from being linked to other such mentions, meaning that they can only be referred back to by pronouns or definite NPs.\footnote{An anonymous reviewer has asked about the sense of `referring' or `referentiality' intended here. The term is unfortunately murky: on the one hand, theoretical literature going back to at least \cite{Reinhart1982} has referred to cognitive models such as `file cards' per entity (see \citealt{Krifka2008} for discussion), which can be evoked by language, but which are empirically hard to define or observe. On the other hand the practical needs of coreference resolution in NLP have often equated referentiality with potential for back-reference. In many partial implementations of `tricky' coreference, this has amounted to including usually ignored expressions if they are referred back to, such as verbal markables for discourse deixis (`she visited... the visit'), or limited markup of coordinations when they are mentioned again with a plural pronoun (`Kim and Yun ... they').} Although indefinite NPs can indicate a variety of semantic or pragmatic functions \citep{BhatiaEtAl2014}, all ``indefinite noun phrases, which begin with the indefinite article (a, an)'' \citep[6]{BBNONEngGuidelines} as well as bare NPs (bare plurals and singulars with no article) are considered generic by definition in ON. Thus the following are \textit{not} annotated in the corpus in any way:

\ex. [Program trading]$_i$ is ``a racket,'' ... [program trading]$_i$ creates deviant swings\\ (ON, wsj\_0121, no coreference in the corpus)\label{ex:indef:trading}

This indirectly implies that coreference resolution systems pairing a pronoun or definite mention with the first NP in such a chain will be penalized, and that systems must learn in an environment in which `it' + `program trading' is once a positive match and once a negative one, despite the fact that it is semantically the same `it', which in fact refers to the generic concept of program trading. The same guideline applies to generic pronouns, as in the following example, which is also left unannotated in ON.

\ex. [you]$_i$ couldn't start unless [you]$_i$ knew that the replacement heart would make it to the operating room \citep[5]{BBNONEngGuidelines}

In this case, systems must again learn that the usual coreference between two instances of `you' in the same sentence does not apply, even though semantically, I would again argue that both pronouns do in fact refer to the same (albeit generic) thing -- anyone who could have started a heart replacement in this context. From a set-theoretical perspective it may even be possible to argue that that set (people who could have started heart replacements at the time) is completely specified, and non-generic (or at least, we could continue the text in a way which explicitly states the names of the relevant people).

The problems with the abstract/generic guideline as a foundation for universal, cross-linguistic coreference annotation are numerous:

\begin{enumerate}
\itemsep0em 
    \item Indefinite mentions are often neither generic nor underspecified or abstract (\textit{``[participants] comprised [15 women and 10 men]''})
    \item It is not immediately obvious that we should not want coreference information even for those mentions which are generic, abstract, etc. (e.g.~\textit{``program trading''}), especially given that we do resolve pronouns referring back to them
    \item In context it is often hard to know whether pronominal mentions are generic, and even when they are, they can still form multiple distinct clusters
    \item Many languages do not have widespread articles which could be used to identify generic mentions, even if we agreed that all indefinites should be considered generic
\end{enumerate}

To illustrate this, we can consider the following examples from the GUM corpus \citep{Zeldes2017}, a freely available English coreference data set with texts from 12 written and spoken text types, which does not exclude indefinite anaphors:\footnote{See the corpus website (\url{https://corpling.uis.georgetown.edu/gum}) for example sources, and the Appendix for longer examples.} 

\ex. Marbles is the first social media star to have [a wax figure]$_i$ displayed in Madame Tussauds ... In 2015 , Marbles unveiled [a wax figure of herself]$_i$ at Madame Tussauds \\(GUM\_bio\_marbles) \label{ex:wax}

\ex. I have [a mini-MMPI]$_i$ ... I have [a chart that I'll go through]$_i$ (GUM\_interview\_dungeon)\label{ex:mmpi}

In \ref{ex:wax}, the wax figure is a specific, unique physical object, mentioned twice as an indefinite NP in the same document. In \ref{ex:mmpi}, the `chart' is referring in context to the exact same entity as the mini-MMPI (a `Minnesota Multiphasic Personality Inventory') -- without the coreference information, a tool or researcher accessing the data automatically would not know that the MMPI was a chart, and vice versa, despite such information sharing being an obvious application of coreference resolution.\footnote{An anonymous reviewer has cautioned that this should not imply that disjoint indefinite mentions should corefer, as in `I always owned [a dog]$_{generic}$. For my 6th birthday, I got [a dog]$_x$. When it died we got [a dog]$_y$ from the animal sanctuary.' I fully agree that there are three distinct `dog' entities in this case, where x and y are subsets of the generic `dog', and therefore not coreferring.}

Generic pronouns too can often serve different purposes and belong to different, meaningful clusters:

\ex. [You]$_1$ feel like [you]$_1$'re prepared , [you]$_1$'re in a , [you]$_2$ know , in a relationship ... [you]$_2$ know (GUM\_vlog\_pregnant)\label{ex:genyou}

In \ref{ex:genyou}, which discusses feeling prepared for a pregnancy, all instances of `you' are arguably generic, but they can be grouped into two sets: the first set refers to some person who might be considering a pregnancy (thereby implicitly ruling out those who cannot become pregnant), while the second set is used exclusively as a pragmatic discourse marker \citep{Oestman1981} to refer to a generic audience and solicit presupposed agreement. It is therefore clear that in context, the person who may be prepared is the same person filling the subject role for `feel', while `you' in `you know' can refer to anyone who may be listening, which forms a different set of individuals (the original utterance comes from a vlog, and the audience is therefore underspecified even for the speaker).

Finally, the issue of cross-linguistic applicability surfaces already in the ON guidelines for Chinese, which cannot refer to indefinite articles as a criterion for identifying genericity, since Chinese does not use articles as in English.\footnote{The same is true of many languages; for example, Slavic language coreference datasets have included generic NPs, which cannot be distinguished based on articles (see e.g.~\citealt{NedoluzhkoEtAl2009} for Czech).} The guidelines instead refer to specific examples:$\!\!\!\!\!\!$\chinese{``死刑} (capital punishment),$\!\!\!$\chinese{世界} (world -- as distinct from ``the world'' meaning the planet Earth),$\!\!$\chinese{社会} (society) are considered generic nouns'' \citep[8]{BBNONZhoGuidelines}. The ON Chinese corpus itself, however, is not consistent about what cases are excluded as generic, and even these specified nouns are regularly annotated as coreferring in contexts which would not be considered definite or specific in English. For `society', these include 44 coreferring instances, as in \ref{ex:zho:soc}.

\ex. \chinese{我们能不能发展的快一些、好一些，实现经济快速发展和  [社会]$_i$  全面进步，并且保持  [社会]$_i$  稳定，十分重要} (ON, cnr\_0016)\\ \textit{W\v{o}men n\'{e}ng b\`{u}n\'{e}ng f\={a}zh\v{a}n de ku\`{a}i y\={i}xi\={e}, h\v{a}o y\={i}xi\={e}, sh\'{i}xi\`{a}n j\={i}ngj\`{i} ku\`{a}is\`{u} f\={a}zh\v{a}n h\'{e} [sh\`{e}hu\`{i}] qu\'{a}nmi\`{a}n j\`{i}nb\`{u}, b\`{i}ngqi\v{e} b\v{a}och\'{i} [sh\`{e}hu\`{i}] w\v{e}nd\`{i}ng, sh\'{i} f\`{e}n zhòngy\`{a}o
} \\ Whether our development can progress faster and better to make it possible for the economy to grow quickly and for [society]$_i$ to make progress across all metrics as well as to maintain the stability of [society]$_i$ is very crucial\label{ex:zho:soc}

Rather than concluding that these cases are annotation errors, I would like to argue below in line with previous work showing the salience and coreferring potential of a range of indefinite phrases (see e.g.~\citealt[295--300]{Kunz2009}) that it is reasonable to annotate them, just as the recurring annotations suggest that annotators regarded them as coreferring.\footnote{One reviewer pointed out that we may still want to use articles in guidelines for languages that do distinguish them, as this can raise inter-annotator agreement, which may be the more important factor for some applications. I certainly do not mean that any reference to articles in guidelines for such languages should be forbidden; but conversely, ruling out `all indefinite anaphors' in coreference datasets a priori, based on form alone, will not produce comparable data across languages, and as I hope this section shows, this would make us lose important information for English as well.}

\subsection{Compound modifiers}

Compound modifiers (or noun-modifying nouns) in English, as in many languages with similar `bare' adnominal modification, often have low referential content, and are known to resist pronominalization (i.e.~they are `anaphoric islands', \citealt{Postal1969}), making them attractive to exclude categorically. For example, as a compound nominal premodifier, it is difficult to imagine pronominal reference to `animals' introduced by a compound `animal hunters' in \ref{ex:animal}. Because of this often limited referential potential, and the desire to avoid disagreements, a conservative annotation scheme may prioritize simplicity and consistency by prohibiting compound modifiers from entering into coreference relations. At the same time, repeated mention of proper nouns in the same position seems normal, and not indefinite or non-referential, as in \ref{ex:hongkong}:

\ex. *[Animal]$_i$ hunters tend to like [them]$_i$. \citep[230]{Postal1969}\label{ex:animal}

\ex. The [Hong Kong]$_i$ government's jurisdiction is the [Hong Kong]$_i$ Special Administrative Region\label{ex:hongkong}

ON makes an exception for proper noun modifiers \citep[3]{BBNONEngGuidelines}, but excludes common noun cases like \ref{ex:market}, which are left unannotated despite subsequent definite reference.

\ex. small investors seem to be adapting to greater [stock market]$_i$ volatility … Glenn Britta … is “factoring” [the market’s]$_i$ volatility “into investment decisions.” (ON, wsj\_0121) \label{ex:market}

However it is by now well known that compound modifiers are not strict islands even to pronominal anaphora \citep{WardEtAl1993}, and corpus data provides plentiful counter examples, leading to many of the same issues as those involved with generics (or more generally, indefinites): noun modifiers can be specific and can be referred back to with definite expressions as in \ref{ex:market}, and even when they are truly generic, we may want to know about their coreference. In \ref{ex:cinammon}, the indication that cinnamon is a spice (and not a brand, a person's name, or something else) is made explicit via coreference, and it is not clear to me why we should exclude such cases from the target gold standard for annotation and NLP tools.

\ex. [Cinnamon]$_i$ basil really does smell like [the sweet spice]$_i$ (GUM\_whow\_basil)\label{ex:cinammon}

As before, the restriction on compound modifiers in English is problematic from a cross-linguistic perspective, and in fact, construct-state modifiers in the ON Arabic coreference annotations, which are very similar to English compound modifiers (e.g.~modifiers resist bearing articles, the translation of \ref{ex:animal} would be ungrammatical), are not restricted in the same way, as seen in \ref{ex:ara} with a semantically specific modifier:

\ex. \<محاكمة ضباط روس بتهمة   الاهمال ... محاكمة  ... لثلاثة ضباط روس >  ~~(ON, ann\_0006)  \\  \textit{mu\underdot{h}\={a}kamatu \underdot{d}ubb\={a}tin r\={u}si bituhmati l-ihm\={a}li...  mu\underdot{h}\={a}kamatun... li-\ubar{t}al\={a}\ubar{t}atin \underdot{d}ubb\={a}tin r\={u}si} \\ \quad[Russian Officer]$_i$ Trial on Charges of Negligence… a trial… for [three Russian officers]$_i$ \label{ex:ara}
\par\vspace{-10pt}\par

Here the Russian officers in the `Russian officer trial' are exactly the same set of `three officers'. Such examples with specific indefinite NPs create the most obvious motivation for including modifier nouns, but the same construction and coreference appear equally with non-specific/generic referents. For parallel corpora and applications of coreference resolution in machine translation too, we can note that what is expressed by such  modifiers in one language may be expressed by a full NP in another, meaning that excluding them will lead to more cross-linguistic divergence (see \citealt[31]{NovakNedoluzhko2015} for example for English and Czech). 

The inconsistency between the annotations of modifier nouns in the three ON languages is thus not only odd from a theoretical perspective; it can cause problems from a practical one as well if we try to develop multilingual applications relying on automatic coreference resolution, only to find out that even systems trained on ON itself behave fundamentally differently for different languages.

\subsection{Nesting} 

Nested coreference, i.e.~a mention with a coreferring mention within its own span, may seem odd, but is in fact rather frequent, and attested in this very sentence. ON does allow nested coreference, but excludes cases without subsequent mention, leading to another difficult learning task for systems (the non-markable status of such cases hinges on later sentences), and unresolved pronouns in text, as in \ref{ex:star}, which remains unannotated in OntoNotes, presumably because outside of the larger phrase's boundaries, it appears to be a singleton.\footnote{In fact, some systems have relied on this constraint, sometimes called `i-within-i' due to the co-index $i$, to rule out certain kinds of match candidates, see \citet[896]{LeeEtAl2013}.}

\ex. [an elusive sheep with a star on [its]$_i$ back]$_i$\label{ex:star} (ON, wsj\_0037)

A further complication in ON concerns nested dates, in which only the top-level expression is considered referential. For example, if a year is mentioned by itself, it may corefer with other mentions, but if it is part of a date, then it may not (again, the example is unannotated in ON):

\ex. what are the opportunities for new developments in the wake of the [1999]$_i$ handover? On December 19, [1999]$_i$, the eve of Macau's transfer of sovereignty... (ON,  ectb\_1001)\label{ex:nested:year}

In \ref{ex:nested:year}, we are unable to discover that the year 1999 in the first sentence is the same year 1999 as in the second, since annotation is ruled out by the ON \textit{form-based} guideline on nested dates. Some readers may think this is not so bad, since we can surely analyze both dates and discover that the year is the same; but I would argue we need to keep in mind that date detection is not trivial (1999 could be a price, or something else), and that these two cases should co-refer based on a \textit{semantic} criterion (they refer to the same thing in the world). 

Moreover, for system development it is problematic to create examples where algorithms will learn that date expressions are atomic, since we can have overt underspecified anaphoric expressions targeting subspans of dates:

\ex. between March 18, [2005]$_i$ and May 7 of [the same year]$_i$ (ON, a2e\_0016)\label{ex:same:year}

Example \ref{ex:same:year} actually \textit{is} annotated in ON, in contradiction with the guidelines, perhaps because of how counterintuitive their effect would be in this case and others like it (this example is by no means unique in ON, and others are left unannotated as intended). Without coreference in \ref{ex:same:year}, we have no chance of discovering that ``2005'' is ``the same year'' in which ``May 7'' takes place. Such information can be crucial for identifying DCT (Document Creation Time, \citealt{RayEtAl2018}) and for Event Timeline extraction (e.g.~for clinical events, \citealt{NikfarjamEtAl2013}), among other applications. However the main point from my perspective is that this behavior is unexpected from a semantic point of view, and is narrowly tied to specific forms, not meanings.

\subsection{Predication}\label{sec:pred}

Judging by discussions with some of my colleagues, perhaps the most controversial opinion I can express in this paper is that the omission of predication from mainstream coreference resolution datasets and systems is a mistake.\footnote{This is not to say that there are not several datasets which mark up coreference for predicates (see below); however since most kinds of predication are not covered by OntoNotes, mainstream papers, systems and the CoNLL shared tasks have not included these cases. For examples of how I think it should be handled, see the Appendix, as well as Section \ref{sec:scopes}} Following an early period in the development of coreference datasets (notably ACE, \citealt{DoddingtonMitchellPrzybockiEtAl2004}) in which most nominal predicates were considered to corefer to their subjects (i.e.~``Kim is a teacher'' $\Rightarrow$ ``Kim'' $\xleftrightarrow[]{coref}$ ``a teacher''), criticism expressed for example by \citet{DeemterKibble2000} and \citet{Zaenen2006} led to ON rejecting coreference in predication altogether. In this section I would like to suggest not only that ON's guidelines are an overreaction to the original criticism, but also that omission is neither a sufficient nor a necessary remedy for the problems that led to its rejection, an outcome left open by \citeauthor{DeemterKibble2000} themselves. 

The core of the problem with predication has been expressed as ``change over time'' (\citealt[11]{HirschmanChinchor1998}, \citealt[632]{DeemterKibble2000}), as in \ref{ex:change:time} (ibid.), but I would argue that it is more broadly a problem of ``change of scope'', since modal scopes create the same type of issue, as in \ref{ex:change:modal} (a made-up example).\footnote{Massimo Poesio has pointed out that the problem is even broader and ultimately stems from the multiple types of entity expressions in names or variables, truth value expressions and quantification in the sense of Montagovian type shifting, see \cite{Partee1986}. In this discussion I will limit myself to the prominent and frequent issues in narrowly defined copula predication, but indeed similar problems arise for negated NPs, relational indefinites and other `tricky NPs' (see \citealt{Landman2004}).}

\ex. [Henry Higgins, who was formerly [sales director of Sudsy Soaps]$_i$]$_i$, became [president of Dreamy Detergents]$_i$\label{ex:change:time}

\ex. If [Beyonc\'{e}]$_i$ were [the Queen of England]$_i$, [she]$_i$ would....\label{ex:change:modal}

In \ref{ex:change:time}, Henry Higgins is not simultaneously sales director at one company and president of another, raising doubts as to the value of a coreference chain including all of those NPs. Likewise in \ref{ex:change:modal}, the pronoun ``she'' arguably corefers more to the Queen of England (in a world where Beyonc\'{e} is the Queen), but not to the first mention of Beyonc\'{e}. In fact, this case verges on examples of non-referential bound anaphora, also discussed by \citeauthor{DeemterKibble2000}, but which cannot be discussed here for space reasons. 

So should we rule out predication due to such problems? After careful consideration of different opinions in the reviews of this paper, I believe the best answer is `yes, predication is different from other types of coreference', but also `no, we cannot ignore it if we want to get the full picture'. My reluctance to ignore predication is based on a double dissociation: not all cases of predication raise this problem, and the problem can arise without predication. The most obvious case for including predication is in identificational predicates, which in English usually involves a definite predicate, but in other languages (e.g.~Japanese) can often be indistinguishable from other predications. Compare these cases, marked up according to ON guidelines, which include naming predicates as copular \citep[27]{BBNONEngGuidelines}:

\ex. [Elizabeth II]$_i$ is the Queen of England. [She]$_i$ ...

\ex. The Queen of England is [Elizabeth II]$_i$. [She]$_i$ ...

\ex. She was crowned [Elizabeth II]$_i$ in 1953. [She]$_i$ ...

Although they forbid annotating subject and predicate as a coreferent pair, ON guidelines do specify a hierarchy for determining which of the two should be taken as the antecedent for subsequent mentions, ranking names above pronouns, and pronouns above definite NPs. These kinds of configurations therefore appear in the official CoNLL coreference shared task dataset and consequently model the output that contemporary coreference resolution systems attempt to produce. I will leave it to readers to imagine scenarios in which the missing member of the chain will lead to loss of information,\footnote{See also \citet[902]{LeeEtAl2013} for an example.} but suffice it to say that including them would not lead to the scope problems above; in fact, some corpora already separate indefinite predication from regular coreference such as ARRAU \citep{PoesioArtstein2008} and GUM, which treat identificational predication as regular coreference, and either label predicative markables whenever semantically applicable (ARRAU), or maintain a special coreference relation subtype for non-identificational cases (GUM, see Appendix).

For the other side of the dissociation, we can consider chains of simple, definite NPs, in which the scope problem does arise, as in \ref{ex:np:scope}.

\ex. [A fresh major in the Swedish army], in 1812 [Gordon] went to war … In 1875 [the now general in the Russian army] was ready to pursue [his] ultimate achievement… [Gordon] is buried in… (adapted from GUM\_bio\_gordon)\label{ex:np:scope}

We can introduce temporally inconsistent mentions (the buried man is not a Swedish major, the Swedish major is not a Russian general), and similarly modality inconsistencies or ambiguous mentions, etc.~Although predication, naming, objects of `as' and other constructions are \textit{likely} locations for this in English, they are not the problem in themselves (so not a necessary condition), and they do not guarantee that there is a problem (so not sufficient). I agree such problems need to be addressed (see Section \ref{sec:solutions}), but with few exceptions, the solution in most major datasets so far has been to largely ignore them, rather than to mark them up as a special case of anaphora (similarly to bridging, discourse deixis or split antecedents). 


\section{Isn't this someone else's job?}

Before suggesting possible solutions, I would like to briefly outline why two potential objections to this paper's proposals do not offer alternative solutions to the problems raised here. Although there may be other suggestions on how to deal with the phenomena not covered in OntoNotes, most probably follow one of two ways of `punting' the issues: either to syntax or semantics.

\subsection{Can't we get all of this from syntax?}\label{sec:syntax}

This line of reasoning is often raised in defence of coreference datasets which exclude predication, but which do have gold treebanking information: since syntax trees represent predication, naming constructions, nesting, and other structures more or less unambiguously, can't we just leave them out of the coreference annotation proper and recover them from the trees?

The answer is no. I have yet to see a convincing case where this has been done, and there are good reasons why we should not think that it is possible. Contrast the following pairs, adapted from GUM:

\ex. \label{ex:would:pair} \a. [He] would be a Libertarian today  \textit{(\textbf{no coref}, since this is hypothetical: he is *not* a Libertarian)}
 \b. [The principles governing an F-E translation]$_i$ would then be: [reproduction of grammatical units; consistency in word usage; and meanings in terms of the source]$_i$ \\ \textit{(\textbf{coref}, since in fact, those \textit{are} the principles of F-E translation)}

 
\ex.\label{ex:substance} 
  \a. [This coffee table] is glass \textit{(\textbf{not predicative coref}, only specifies the substance that most/part of the table is made of)}
 \b. [This ice here]$_i$ of course is [water]$_i$ \textit{(\textbf{predicative coref}, part of a chemistry demonstration in which the speaker literally identifies an ice cube as being the same water in a solidified state)}

\ex.\label{ex:doctor} 
  [He]$_i$ was not the leaf-collecting doctor, but [an altogether strange man, with silver eyebrows in his smooth face and long fine-knuckled hands]$_i$ (GUM\_fiction\_lunre)

In \ref{ex:would:pair}--\ref{ex:substance} the same syntactic construction yields coreferring expressions in one case and no coreference in the other (in \ref{ex:would:pair}, identity coreference, in \ref{ex:substance}, predication coreference, labeled as such in GUM). In \ref{ex:doctor}, only part of a coordinate predicate NP is coreferring, meaning that extracting the correct span using only syntax is non-trivial, and of course the phenomena in these examples can co-occur (imagine processing modal coordinate substance predications correctly!). Syntactic approaches also assume a well-formed syntax tree uttered contiguously by a single speaker, which in some cases is not a given.

It is also extremely difficult to know whether fairly mundane spatio-temporal predicate NPs are coreferring, regardless of definiteness:

\ex. This town is 35 minutes from the harbor \textit{(\textbf{no coref}, purely spatio-temporal `is', since town$\neq$35 minutes)}

\ex. But Christmas is still the whole winter to wait \textit{(\textbf{no coref}, but most syntax trees would show `winter' as a definite predicate NP)}

Similarly for NP-internal or nested coreference resolution, we could easily have ambiguities in OntoNotes-style data. Consider this minor modification of the `sheep' example above

\ex. A sheep$_i$ in a coat$_j$ with a star on its$_{i/j}$ back\label{ex:coat}

The data in \ref{ex:coat} means that nested resolution is needed for disambiguation and cannot be extracted from syntax automatically, just as it is for predication. Finally for compound modifiers, although many coreferential cases include verbatim repetition of an entire compound, which can perhaps be recognized from word forms alone, some cases do not, with examples like \ref{ex:cinammon} above forming the most striking cases. There can be no syntactic solution if we want to recover such relations.

To be clear, this is not to say that morpho-syntactic criteria can never be used in coreference annotation guidelines in any way: the reason why projects have used syntactic constructions diagnostically is because they can be very helpful in differentiating subtly different constructions and as a result help raise inter-annotator agreement. The point is rather that use of such criteria should be in service of semantic distinctions, which should be our actual object of interest: syntax and morphology should not be allowed to rule out things which we are certain do co-refer (for example morphologically incongruent singular and plural NPs), and they should not be used to admit things that do not (for example appositions which do not corefer). And if something is supposed to be recoverable from syntax, this should not move us to exclude it from coreference annotation -- otherwise we are effectively mandating that any coreference resolution system will also have to tackle syntactic parsing.

\subsection{How about semantics?}

Another line of reasoning is that for `quirky' coreference cases, including predication but also various kinds of distributive semantics (which space prevents me from discussing), semantic annotation should be in charge of annotating predicate structure in a way that disambiguates the issues in Section \ref{sec:syntax}. The problem here is that even the most elaborate semantic annotation formalisms available right now do not address the main problems, since predication is largely subsumed in lexical semantics. In PropBank \citep{PalmerGildeaKingsbury2005}, one of the simpler but most widespread types of semantic annotation, predicates are simply argument structure graphs with word sense disambiguation, but copula `be' is simply annotated as \textsc{be.01}, as in the following example from OntoNotes:

\ex. \a. the most special is rice (ON, cctv\_0000)
\b. Prop: \textsc{be.01}, ARG1: `the most special', ARG2: `rice'

It is therefore impossible to know whether this is a coreferential case or not (it is, specifically a non-identification predicative coreference: in this case a rice plant is being discussed). 

More complex semantic analysis is undertaken in the less widespread but highly detailed formalism of Abstract Meaning Representation (AMR, \citealt{BanarescuEtAl2013}), which has facilities to indicate, for example, predication negation, possibly solving the problem in \ref{ex:doctor}; but AMR is not aligned to words, and it is impossible to use it for the purpose of coreference resolution, as the following example from the Little Prince corpus (ibid.) illustrates:

\ex. \a. It is my fault that you have not known it all the while. 
    \b. (f / fault-01  \\  $~~~~~~$:ARG1 (i3 / i) \\
     $~~~~~~$:ARG2 (k / know-01 :polarity - \\
     $~~~~~~~~~~~~$ :ARG0 (y / you) \\
             $~~~~~~~~~~~~$ :ARG1 (i2 / it) \\ 
             $~~~~~~~~~~~~$ :time (w / while-away-01 \\
                 $~~~~~~~~~~~~~~~~~~$:duration (a / all)))) 

The AMR analysis captures the argument structure of `fault' perfectly (arguments: the spearker and the `knowing' predicate), but leaves no indication of the original expletive subject `it', its relationship with the extraposed clause `that you have not known...', etc. Because of AMR's (intentional) distance from word forms in the original sentence, it does not fulfill or replace the task of text-anchored coreference resolution -- AMR simply has no obligation to include nodes for each surface pronoun or NP. Similarly, neither of these formalisms analyzes the properties of discourse referents in terms of genericity, or contemporaneous temporal or modal scope.

\section{Fixing the problems}\label{sec:solutions}

Is it impossible to envision a semantic or syntactic formalism that can represent the facts discussed here? Certainly not: we could just extend semantic or syntactic annotation to distinguish such cases. But here I would like to argue that covering these cases is part of the job of coreference resolution, and realistically, if coreference resolution does not tackle them, they are unlikely to be covered by other annotation efforts in the current landscape. 

\subsection{Unexiling tricky coreference}

Although it should be clear by now that I would like to see a lot more things annotated in coreference resolution datasets, I would like to emphasize that I am not saying that all of these types of coreference are \textit{the same}. Addressing them does not necessarily mean that we need to lose the distinction between predication, pronominal anaphora and other types of coreference. OntoNotes itself distinguishes several types of apposition from identity coreference, and several other corpora distinguish not only appositions, but also pronominal anaphora from lexical identity (e.g.~GUM) or non-identificational predication (e.g.~both GUM and ARRAU for English, or CESS-ECE for Spanish, \citealt{RecasensEtAl2007}, which are all useful examples of explicitly handling, rather than ignoring predication), in addition to harder phenomena such as discourse deixis or bridging anaphora. 

Critics may point out that including trickier cases will inevitably lead to lower agreement, but I would answer that doing so should not decrease agreement on `easy cases', and that in instances of disagreement, most often each of the conflicting analyses corresponds to a valid reading of an ambiguous context (as demonstrated by \citealt{PoesioArtstein2005}). Human annotators also tend to disregard or not notice a relation more often than connecting non-coreferring entities \citep{JiangEtAl2013,Zeldes2017}, meaning data would still have relatively high precision (for GUM, which includes the phenomena discussed here, student annotators achieved precision, recall and F1 of 0.918/0.811/0.861 compared to the adjudicated gold standard, when minimal mention span matching is used, \citealt[596]{Zeldes2017}). With this in mind, including these phenomena and labeling them as such, as is done in ARRAU or GUM, seems in essence already close to the ideal solution, with a few caveats and additions.

The first is that the default coreference resolution task should, in my opinion, cover all types of cluster-based\footnote{I use this term to set aside other types of anaphora, such as bridging, sense anaphora (e.g.~`one' anaphora), etc., not because they are not important, but because we already have much work to do on many types of identity coreference and closely related concepts.} coreference discussed here, including compound modifiers, indefinites and predication. As shown in Table \ref{tab:freqs}, which tallies types of anaphoric expressions in GUM, all of these are extremely frequent, non-marginal phenomena, which often overlap with clear cases of identity coreference (e.g.~definite copula predicates, semantically specific compound modifiers). Taken together, indefinite NPs, compound modifiers and predicates form around 19\% of all previously mentioned NPs, occurring about 26 times per 1K tokens. For comparison, this is more than the remaining definite common NPs (21.67 times) or subsequently mentioned proper names (19.84). Nested dates (not shown in the table) are also ubiquitous in GUM data, with 13.8\% of temporal NPs being nested in other temporal NPs (including singletons) and 38\% of coreferring year expressions in the corpus being nested in a date entity span, which would exclude them in ON. Removing such mentions from the target standard for mainstream NLP leads to counterintuitive results and ultimately creates an incongruity between systems' performance in shared tasks and their coverage for real world applications, which often assume the premise of being able to collect \textit{all} information about an entity in a text (say, some year) when coreference resolution is `correct'.

\begin{table*}[t!b]
\centering
\resizebox{0.5\textwidth}{!}{
\begin{tabular}{lrr}
\toprule
\textbf{expression type} & \textbf{instances per 1K tokens} & \textbf{\% of total} \\
\hline
\textit{pron. anaphora} & 59.84 & 44.98\% \\
\textit{cataphora} & 1.39 & 1.05\% \\
\textit{nominal predicate} & 4.94 & 3.71\% \\
\textit{compound mod.} & 5.79 & 4.36\% \\
\textit{split antecedent} & 1.22 & 0.92\% \\
\textit{apposition} & 3.81 & 2.86\% \\
\textit{coref. name} & 19.84 & 14.92\% \\
\textit{other indef. NP} & 14.53 & 10.92\% \\
\textit{other coref} & 21.67 & 16.29\% \\
\hline
\textbf{total} & 133.02 & 100\% \\
\bottomrule
\end{tabular}%
}
\caption{Coreference link type distribution in GUM}
\label{tab:freqs}
\end{table*}

This position, if accepted, raises the question of what we should do about Henry Higgins, president of Dreamy Detergents? Haven't we been here before, and didn't the criticism of the ACE dataset teach us that `change over time' makes text-anchored coreference futile?

\subsection{Addressing Scope}\label{sec:scopes}

The `change over time', or better, `change of scope' problem raised by \citet{DeemterKibble2000} remains the biggest challenge to a more inclusive view of coreference, and I have answers for it on several levels. On the practical level, scope problems are both not very common and ubiquitous. They are not very common since true hard cases, such as texts covering an entity changing substantially vis-a-vis the relevance of referring expressions over time, are only a subset of texts, and even in those texts, any problematic entity often co-exists with scores of other predicate NPs which do not raise such problems. Refusing to include those hundreds of identificational predicate NPs is throwing out the baby with the bath water.

At the same time, the problem is ubiquitous (cf.~\citealt{RecasensEtAl2012} on `near-identity coreference'). Any long text will involve changes to participants, which, even if they are not stated, mean that not all predicates stated of some entity actually apply simultaneously. If we omit predicates from the coreference annotation and then tell the story of Henry Higgins, he does not remain the same Henry Higgins whether we include the information about his various jobs or not, or whether we express that information using NPs or not -- we could just say he was `fired' and alter the properties of the entity (i.e.~his job) without interacting with coreference. To then say that we will use syntax or semantics to describe Henry's biography does not absolve us of the need to address scope, since the same inconsistencies will result in a semantic annotation of the text. But ignoring `difficult' predicate NPs as part of coreference annotation will lead to many gaps in what practitioners using coreference resolution might expect or could harness.

Moving over to the theoretical perspective, I think there is enough merit in representing scope that we as a community of researchers interested in coreference should consider what is the right thing to do. The fact that we recognize the problem with Henry Higgins suggests that we \textit{do} know that both the `sales director' Henry and the `president' Henry are the same person: just not at the same time. It could be tempting to simply allow a co-indexed scope attribute in our coreference annotation schemes, as in \ref{ex:beyonce:scope} for the Beyonc\'{e} example.

\ex. If $<$coref id=\textquotedbl1\textquotedbl$>$Beyonc\'{e}$<$/coref$>$ were $<$coref id=\textquotedbl1\textquotedbl{ }scope=\textquotedbl{s}1\textquotedbl$>$the Queen of England $<$/coref$>$, $<$coref id=\textquotedbl1\textquotedbl{ } scope=\textquotedbl{s}1\textquotedbl$>$she$<$/coref$>$ would conduct the annual swan upping.\label{ex:beyonce:scope}

This notation would accurately show that the pronoun `she' refers to its intuitive antecedent entity in the document context, `Beyonc\'{e}', while also indicating that this specific `she' is referential in the world in which Beyonc\'{e} is the Queen, and co-refers to that Queen. We could also add a coreference type indicating that we are looking at identity-predication, etc.

For modal cases, adding a scope ID may be enough, but for temporal ones, we could even consider annotating intervals where these are known (or place-holders where they are not; see the RED corpus for a similar strategy in event annotation, \citealt{OGormanEtAl2016}):

\ex. $<$coref id=\textquotedbl1\textquotedbl$>$Henry Higgings, who was formerly $<$coref id=\textquotedbl1\textquotedbl{ }scope=\textquotedbl{s}1\textquotedbl{ } from=\textquotedbl1999\textquotedbl{ }to=\textquotedbl2001\textquotedbl$>$sales director at Sudsy Sopes$<$/coref$><$/coref$>$, became $<$coref id=\textquotedbl1\textquotedbl$>$ President of Dreamy Detergents$<$/coref$>$

Omitting scope IDs and intervals would mean that these are the `general' scope conditions of the document, and that further information is not known.

Alternatively, given that there have been independent efforts to annotate modal \citep{hendrickx-etal-2012-modality,nissim-etal-2013-cross,rubinstein-etal-2013-toward} and temporal scope \citep{pustejovsky-stubbs-2011-increasing,styler-iv-etal-2014-temporal,PustejovskyEtAl2019} for semantics, it may make sense to encode scope independently of coreference annotation, as a span level property of sentences, or VPs or any text span, which interacts with coreference, since many other types of annotation could be affected by modal and temporal scopes. For example, the hypothetical `conduct the swan upping' (an annual census of the Queen's swans) in \ref{ex:beyonce:scope} is also restricted to the modal scope in which Beyonc\'{e} is the Queen, and event annotations of this predicate could take this into account.

I do not want to pretend that annotating scope for coreference is simple, or that we are in a position to easily add it to large benchmark datasets such as OntoNotes: there are surely many complications, even without which this would mean much work, and a detailed examination of how coreference interacts with existing scope annotation schemes is certainly in order. But assuming we can take the position that (serious) scope problems are fairly rare, then I think the right starting point before moving on to an adequate representation for these cases is to include the phenomena missing in OntoNotes as an ideal target, and offer scope as a topic for further advanced research on coreference, similarly to bridging anaphora or split antecedents, whose absence in benchmarks like OntoNotes is well understood, but does not interfere with progress on the most common types of coreferentiality in corpora.

\section{Conclusion}

If this opinion piece falls short of changing any coreference annotation practices, then I hope it at least serves one purpose: to make researchers aware of the limitations of ON-style coreference, and by extension most NLP tools for coreference resolution. My experience has been that these limitations are not well understood by proficient NLP practitioners who use tools such as vanilla e2e coref \citep{LeeHeLewisEtAl2017}, AllenNLP \citep{GardnerGrusNeumannEtAl2018} or Spacy's Huggingface \citep{ClarkManning2016} implementations, and this often leads to surprises. Conscious choices for or against including some of the trickier phenomena will inevitably involve factors relating to trade-offs of linguistic adequacy versus reliability and application domains, but informed decisions are better than ones made spontaneously or based on inertia.

A more ambitious goal for this paper is that readers involved in the production and (re)annotation of coreference datasets will consider following corpora such as GUM and ARRAU in taking a broader view of the phenomenon, and use subtypes to carve out areas which some systems might want to leave out or label in special ways for downstream applications -- the Appendix offers some worked examples from the GUM corpus as a starting point. Including more coreference phenomena does not mean that we have to be naive about the potentially lower level of agreement on complex cases (e.g.~generics, predication); but removing such cases from output that contains them is generally much easier than trying to add them to data in which they are not included in any way, and if modifying data along different lines benefits specific subsets of downstream tasks, then the flexibility offered by distinguishing but including tricky phenomena is all the more valuable. This has also been one of the main lessons I have learned from working on multilayer corpora (see \citealt{Zeldes2018}): no, we cannot get everything from syntax as pointed out above, but syntax is great for finding specific phenomena or manipulating data into different schemes (e.g.~removing compound modifiers, changing how appositions are handled). In one paper, this has even allowed transforming GUM, the dataset most closely matching this proposal, to follow the OntoNotes guidelines (the resulting dataset is called `OntoGUM', see \citealt{zhu-etal-2021-anatomy} for details and evaluation).

Finally, I think that as corpus resources converge across languages on uniform standards, as evidenced in the Universal Dependencies project for treebanking, POS tagging and morphological annotation \citep{MarneffeEtAl2021}, or multilingual initiatives in other areas of discourse (e.g.~the DISRPT shared tasks on discourse relations, \citealt{ZeldesDasMazieroEtAl2019,ZeldesEtAl2021}) we will need to make coreference more consistent across datasets. This will mean that we have to minimize the use of guidelines based on language-specific forms, such as definiteness or specific constructions. Instead, I have argued here that coreference should primarily be decided on semantic grounds, which seem likelier to transfer between languages. As an added motivation, I offer the consideration that cross-linguistically and semantically grounded coreference annotations are likely to work better for applications involving multiple corpora, such as multilingual and cross-document coreference, as well as entity linking (see \citealt{LinZeldes2021} for more on merging entity linking with cross-document coreference).

The problems discussed in this paper are in my opinion not intrinsically ones of copula sentences or modal verbs or other constructions, but rather issues at the discourse-semantics interface, where meanings from individual sentences coalesce to form narratives in which entities change. This suggests that tackling the issues in a cross-linguistically applicable way will ultimately require the development of standards for handling spatiotemporal, modal and other types of scope independently of language or text type.


\bibliography{bibl.bib}

\appendix
\par\vspace{-5pt}\par
\section{Worked examples}

This section gives detailed examples taken from different genres in the GUM corpus, which includes two versions of coreference annotations: GUM's native, exhaustive coreference scheme, which follows the suggestions in this paper, and an automatically produced version of the same data following the OntoNotes guidelines, a rendition of the corpus referred to as OntoGUM (see \citealt{zhu-etal-2021-anatomy} for details). Markup for all examples in this appendix can be found at the GUM corpus repository, in both the original and OntoNotes scheme, at \url{https://github.com/amir-zeldes/gum}.

\par\vspace{-5pt}\par

\subsection{How to Grow Basil}

\begin{figure*}[h!tb]
\centering
\includegraphics[width={0.8\textwidth}, trim={0cm 2.5cm 0.5cm 1.6cm}, clip]{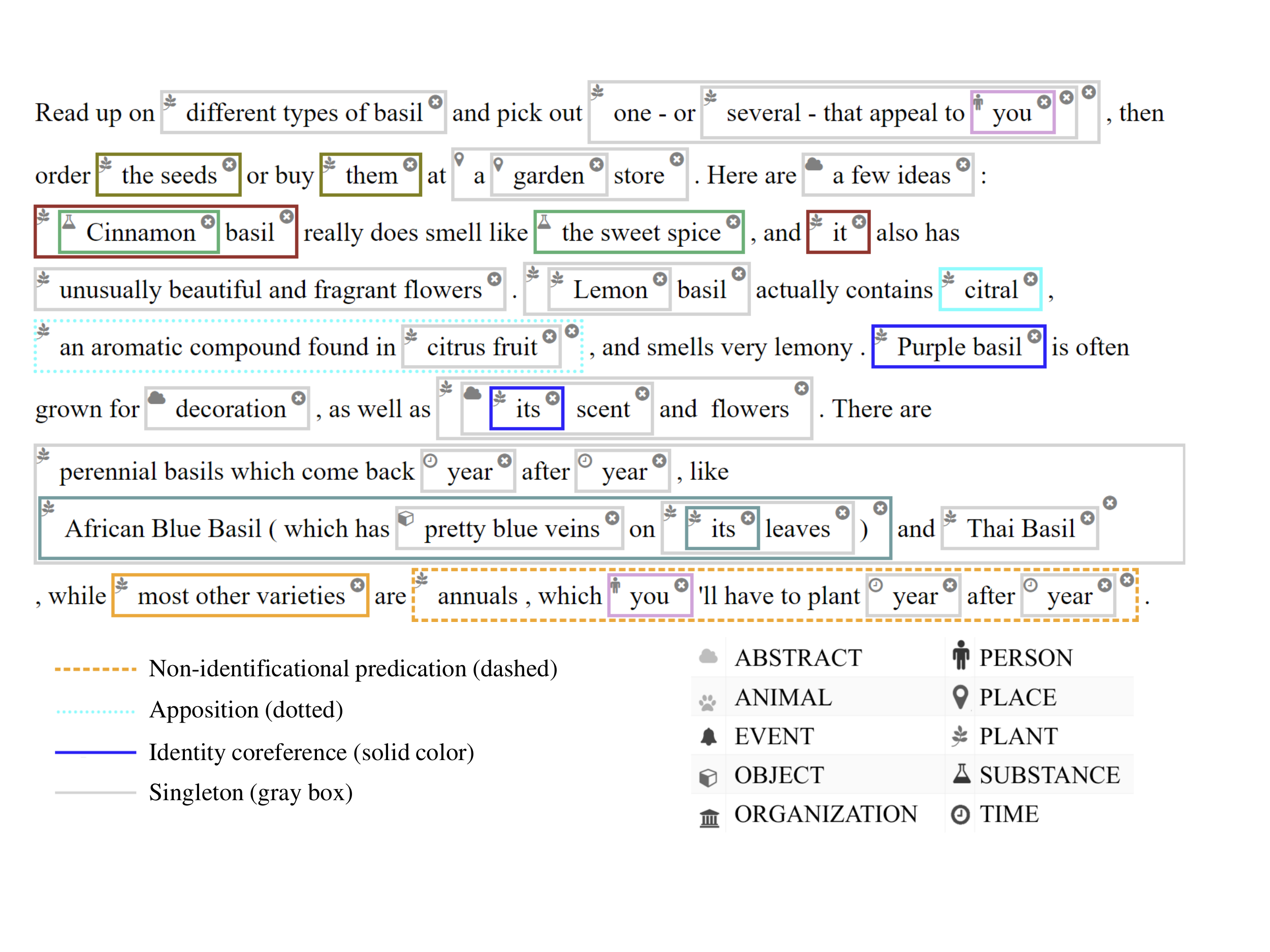}
\par\vspace{-10pt}\par
\caption{GUM-style coreference annotations (GUM\_whow\_basil), visualized using GitDox \citep{ZhangZeldes2017}.}
\label{fig:basil-gum}
\par\vspace{-5pt}\par
\end{figure*}

This excerpt from a Wikihow guide on how to grow basil (see the corpus website for source and licensing information on all texts) illustrates the proposal's coverage for nested identity coreference and non-identificational predicative coreference, as well as singletons and entity type annotations found in GUM. The more sparse OntoGUM representation illustrates the comparative loss of information based on the OntoNotes scheme. 

\begin{figure*}[h!tb]
\centering
\includegraphics[width=0.8\textwidth, trim={0cm 5.5cm 0.5cm 5.6cm}, clip]{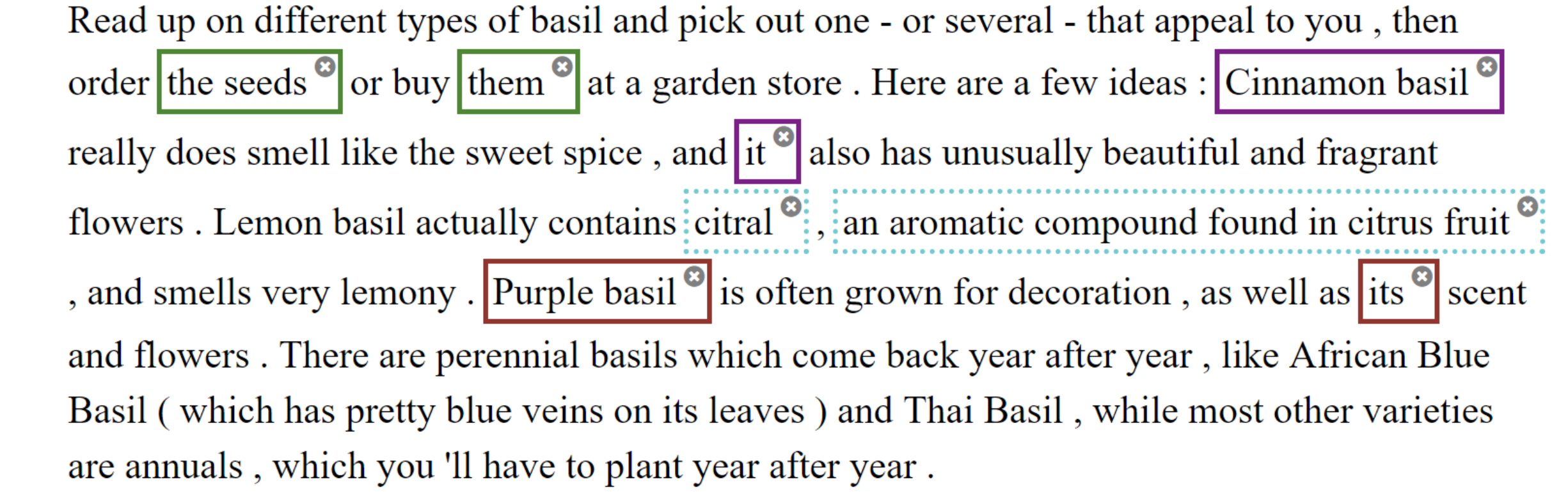}
\par\vspace{-10pt}\par
\caption{OntoNotes-style coreference annotations (GUM\_whow\_basil).}
\label{fig:basil-on}
\par\vspace{-15pt}\par
\end{figure*}

Beyond the singletons (in grey) and entity types (icons in top left corners of boxes) found in Figure \ref{fig:basil-gum} (see the legend), some items to note include the common noun compound modifier antecedent `Cinammon',  the predication between `most other varieties' and `annuals...', and the generic pronoun `you' (in pink). Additionally, we see nested coreference in `[African Blue Basil ( which has pretty blue veins on [its]$_i$ leaves)]$_i$'.  In Figure \ref{fig:basil-on} we see that the nested coreference set is not included according to the OntoNotes scheme, since it is not mentioned a subsequent time outside the nesting mention, and is therefore considered a singleton (and singletons are generally omitted). Similarly the mention of `Cinammon' is ignored, and predication for `annuals' cannot be recovered trivially (see Section \ref{sec:pred}).

\subsection{Spoken indefinites}

This example comes from a conversation (GUM\_conversation\_lambada), originally from the UCSB corpus of Spoken American English \citep{BoisEtAl20002005}, and included in GUM's conversation genre.

\begin{figure*}[h!tb]
\centering
\includegraphics[width=0.8\textwidth, trim={0cm 3cm 0.5cm 2.5cm}, clip]{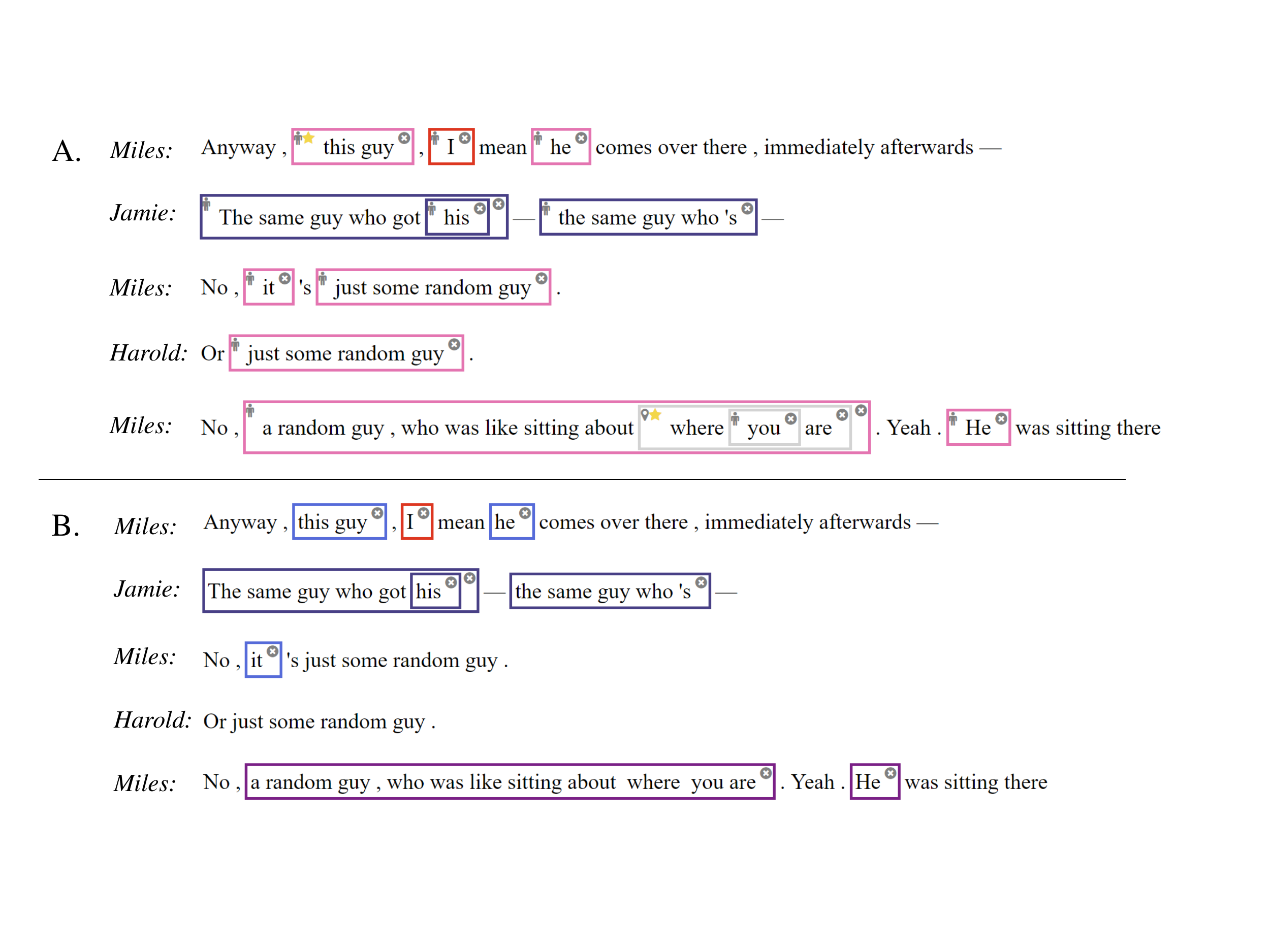}
\par\vspace{-10pt}\par
\caption{Analyses of a conversation fragment according to coreference schemes based on GUM (above) and OntoNotes (below). Referring expressions with a gold star are annotated as accessible in context.}
\label{fig:random}
\par\vspace{-20pt}\par
\end{figure*}

In Figure \ref{fig:random}A, mentions of the `random guy' are all grouped together, regardless of definiteness or predication, since these are not criteria for coreference in the GUM scheme, which follows the proposal in this paper. The OntoNotes version ignores some mentions (if they are indefinites in chain-medial position) and oddly splits the chain into two sets, creating the impression that there are two random guys, in addition to the person being asked about by the second speaker (there are only two guys in context). GUM's mention annotation scheme, which marks up information status (with six subtypes), additionally flags referring expressions as accessible in context, including an initial deictic realization of `this guy' and the situationally accessible `where you are', which are marked by gold stars in the figure. 

\subsection{Datelines and headlines}

This example comes from the news genre (taken from Wikinews) and illustrates two common problems in newswire data.

\begin{figure*}[h!tb]
\centering
\includegraphics[width={0.8\textwidth}, trim={0cm 4.5cm 0.5cm 2.6cm}, clip]{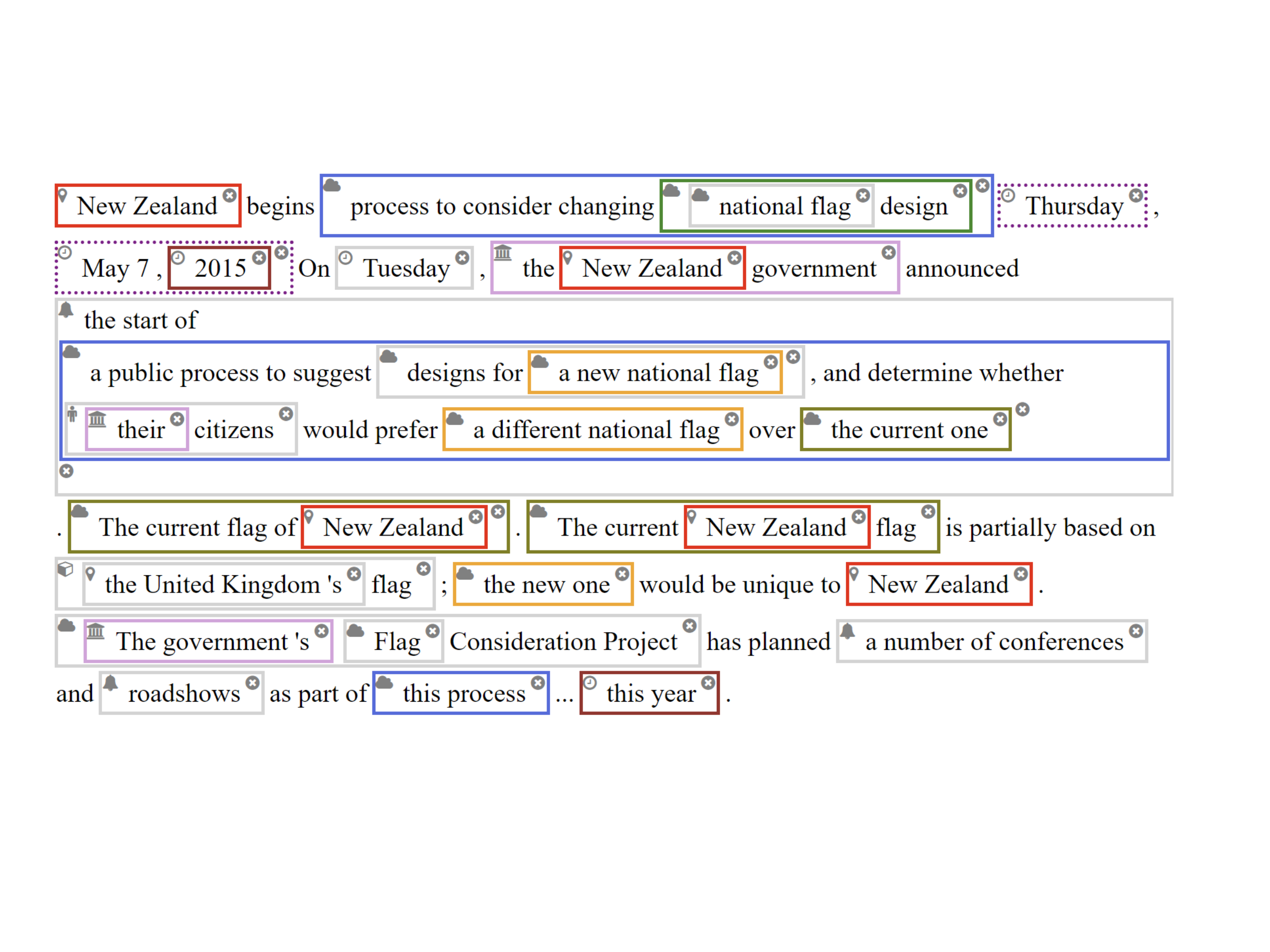}
\par\vspace{-10pt}\par
\caption{GUM-style coreference annotations (GUM\_news\_flag).}
\label{fig:newzealand-gum}
\par\vspace{-5pt}\par
\end{figure*}

\begin{figure*}[h!tb]
\centering
\includegraphics[width={0.8\textwidth}, trim={0cm 5.5cm 0.5cm 4.6cm}, clip]{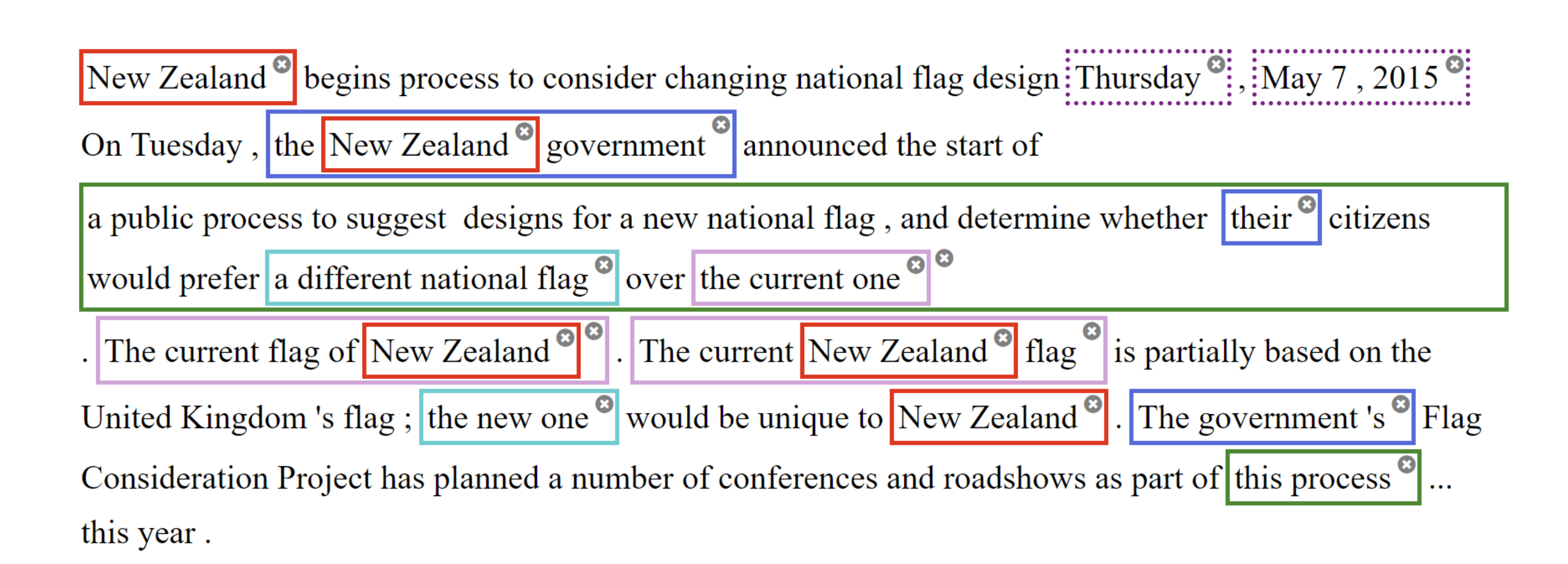}
\par\vspace{-10pt}\par
\caption{OntoNotes-style coreference annotations (GUM\_news\_flag).}
\label{fig:newzealand-on}
\par\vspace{-5pt}\par
\end{figure*}

In Figure \ref{fig:newzealand-gum}, `2015' corefers with `this year', and `this process' corefers with both spans `a public process' and the `process to consider...' which is part of the headline. Because the headline and first subsequent mention of the `process' are both indefinite, the first mention is eliminated in the OntoNotes version in Figure \ref{fig:newzealand-on}. Additionally, because the year in the dateline is nested in a date, OntoNotes guidelines rule it out as an antecedent for `this year', meaning that the latter becomes a singleton, and both mentions are excluded.

\end{document}